\documentclass[letterpaper, 10 pt, conference]{IEEEtran}  
\usepackage{geometry}
\IEEEoverridecommandlockouts                              

\usepackage[numbers]{natbib}
\usepackage{multicol}
\usepackage[bookmarks=true]{hyperref}

\usepackage{subcaption}
\usepackage{units}
\usepackage{amsmath}
\usepackage{amssymb}
\usepackage{tikz}
\usepackage{tikzsymbols}
\usetikzlibrary{shapes}

\usepackage{algorithm}
\usepackage{algorithmic}
\usepackage{multicol}

\usepackage{titling}

\DeclareUnicodeCharacter{2217}{*}

\title{\vspace{0.49cm}\LARGE \bf
A Time-dependent Risk-aware distributed Multi-Agent Path Finder based on A*
}

\author{Samuel Nordstr{\"o}m$^{1}$, Yifan Bai$^{2}$, Bj{\"o}rn Lindquist$^{3}$ and George Nikolakopoulos$^{4}$
\thanks{*This work was not supported by any organization}
\thanks{$^{1}$Samuel Nordstr{\"om} is Robotics \& AI Team in Department of Computer, Electrical and Space at Lule\r{a} University of Technology in Sweden and are the corresponding author
        {\tt\small samuel.nordstrom@ltu.se}}%
\thanks{$^{2}$Yifan Bai is Robotics \& AI Team in Department of Computer, Electrical and Space at Lule\r{a} University of Technology in Sweden
        {\tt\small yifan.bai@ltu.se}}%
\thanks{$^{3}$Bj{\"o}rn Lindqvist is Robotics \& AI Team in Department of Computer, Electrical and Space at Lule\r{a} University of Technology in Sweden
        {\tt\small bjorn.lindqvist@ltu.se}}%
\thanks{$^{4}$George Nikolakopoulos is Robotics \& AI Team in Department of Computer, Electrical and Space at Lule\r{a} University of Technology in Sweden
        {\tt\small  george.nikolakopoulos@ltu.se}}
}

\begin{document}

\maketitle
\thispagestyle{empty}
\pagestyle{empty}

\begin{abstract}
Multi-Agent Path-Finding (MAPF) focuses on the collaborative planning of paths for multiple agents within shared spaces, aiming for collision-free navigation. Conventional planning methods often overlook the presence of other agents, which can result in conflicts. In response, this article introduces the A$^*_+$T algorithm, a distributed approach that improves coordination among agents by anticipating their positions based on their movement speeds. The algorithm also considers dynamic obstacles, assessing potential collisions with respect to observed speeds and trajectories, thereby facilitating collision-free path planning in environments populated by other agents and moving objects. It incorporates a risk layer surrounding both dynamic and static entities, enhancing its utility in real-world applications. Each agent functions autonomously while being mindful of the paths chosen by others, effectively addressing the complexities inherent in multi-agent situations. The performance of A$^*_+$T has been rigorously tested in the Gazebo simulation environment and benchmarked against established approaches such as CBS, ECBS, and SIPP. Furthermore, the algorithm has shown competence in single-agent experiments, with results demonstrating its effectiveness in managing dynamic obstacles and affirming its practical relevance across various scenarios.
\end{abstract}


\section{Introduction}\label{sec1}

Robotic agents are increasingly used as technology advances, and their benefits grow when multiple agents collaborate in shared environments~\cite{lindqvist2022multimodality}. The DARPA Subterranean Challenge~\cite{subt} has boosted multi-agent development, enhancing robotics applications alongside humans and various dynamic obstacles. This novel approach focuses on adeptly avoiding collisions with dynamic objects, relying on knowledge of their size, trajectory, and velocity.

Path planning in robotics has seen significant research, especially concerning dynamic environments~\cite{karur2021survey}. Effective navigation often employs temporal logic, as seen in the T$^*$ path planner~\cite{khalidi2020t}. In contrast, D$*_+$ planners~\cite{karlsson2022d+} create paths for safe navigation, typically assuming static maps or relying on frequent updates, but they neglect future obstacle positions.

Additionally, methods like DTAA~\cite{nordstrom2024dtaa} focus on dynamic obstacle avoidance. However, these systems may struggle to stay on path, occasionally leading to significant detours or collisions with walls due to limited environmental awareness.

\begin{figure}
    \centering
    \includegraphics[width=0.6\columnwidth]{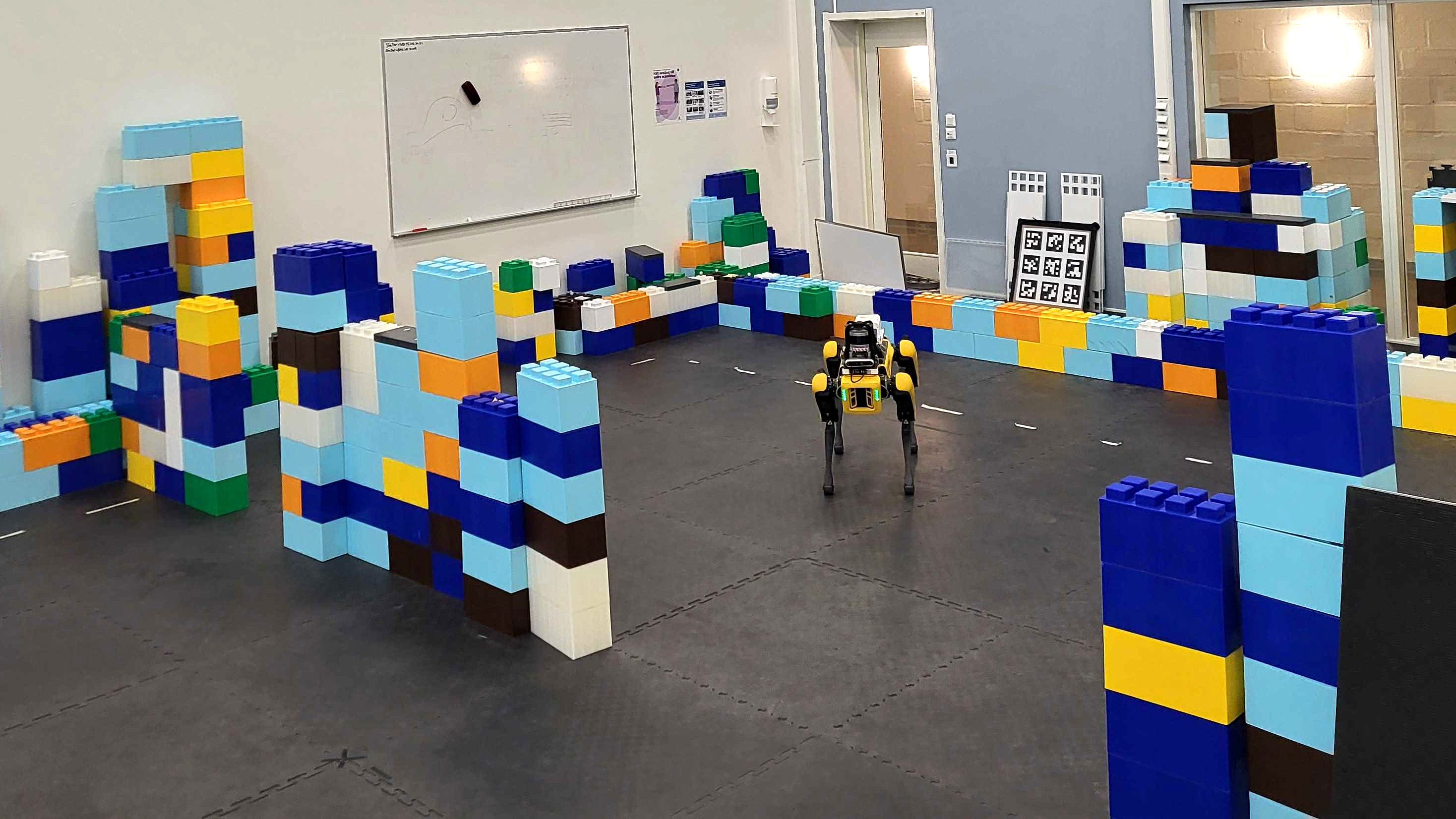}
    \caption{A snapshot from lab experiments where a Spot robot plans a path through a gap in a wall. This is the same experiment as in Figure~\ref{fig:risk:layer}}
    \label{fig:lab}
\end{figure}

In distributed solutions, the ego-agent relies on other agents to provide their motion plans, including size, path, and velocity. For dynamic objects or rogue agents, detection and path prediction processes are essential. Various prediction solutions leverage context and semantic information to enhance accuracy~\cite{rudenko2020human,bartoli2018context, gorlo2024long}. In the absence of such information, a constant velocity prediction model can suffice over short time frames~\cite{scholler2020constant}. One effective path prediction approach is creating a heat map to forecast future positions~\cite{mangalam2020pecnet}.

The Multi-Agent Path Finding (MAPF) problem, defined in existing literature~\cite{stern2019multi}, involves robots occupying vertices in a graph, moving from vertex to vertex or remaining stationary. A MAPF problem is solved when robots can reach their goals without violating movement constraints: \textbf{Vertex conflict}: A vertex conflict occurs when two or more agents attempt to occupy the same vertex at the same time. \textbf{Edge conflict}: An edge conflict happens when two agents plan to traverse the same edge in opposite directions at the same time. \textbf{Follow conflict}: A follow conflict arises when one agent is forced to follow another agent closely in the same direction, leading to a situation where the follower cannot progress or gets stuck. \textbf{Cyclic conflict}: A cyclic conflict occurs when agents are involved in a situation where they continuously block each other in a cycle, preventing progress for any agent involved in the cycle. and \textbf{Swap conflict}: A swap conflict happens when two agents want to swap places, which results in both agents blocking each other, preventing either from moving. This classical definition meets theoretical needs but struggles in real-world scenarios due to positioning inaccuracies, speed variations, and robot size mismatches. While static obstacles can be managed through inflation~\cite{LI2018275}, dynamic obstacles complicate matters. Larger vertex sizes and extending conflict zones can help. Despite limitations, centralized planners like Conflict Based Search (CBS)~\cite{sharon2015conflict}, Enhanced Conflict Based Search (ECBS)~\cite{barer2014suboptimal}, and Safe Interval Path Planning (SIPP)~\cite{phillips2011sipp} have emerged. Distributed approaches also exist, focusing on specific sub-cases~\cite{lee2024dmvc}, and formation control solutions can be centralized or decentralized~\cite{liu2024survey}. Understanding robot navigability is crucial, leading to the development of traversability mappers~\cite{benrabah2024review}. Research employing a next-best view approach incorporates collision risk to yield rapid solutions in high-speed scenarios~\cite{liu2024uncertaintyriskawareactiveview}. Classifying ground traversability into multiple risk levels enhances accuracy~\cite{cai2023probabilistictraversabilitymodelriskaware}. The work in~\cite{10750397}, which converts 3D occupancy maps to 2D traversability maps, is utilized here due to its open-source implementation and compatibility with the planned sensors for the experimental robot.

\subsection{Contributions}
This work establishes a novel distributed multi-agent path planning method that incorporates dynamic obstacle avoidance. In this context, the paths of the agents are shared, and other agents are treated as dynamic obstacles. The proposed approach, A$^*_+$T, utilizes a risk strategic framework similar to D$^*_+$, incorporating a time-dependent component to manage dynamic obstacles as temporal risks during the planning process.

The effectiveness of this solution has been evaluated through multi-agent simulations conducted in Gazebo, where it was compared with various Multi-Agent Path Finding (MAPF) solutions, including Conflict-Based Search (CBS), Enhanced Conflict-Based Search (ECBS), and SIPP. Additionally, the solution has been validated through real-world single-robot experiments, demonstrating that A$^*_+$T can successfully address dynamic obstacles beyond just the trajectories of other agents.

\section{Method}

The proposed path planner, A$^*_+$T, includes key components described in subsequent sections. Section~\ref{sec:A} introduces the main algorithm, while Section~\ref{sec:prox} covers the risk layer that generates a risk map for path traversability. To improve navigation around dynamic obstacles, a temporal dimension is added in Section~\ref{sec:t}. Additionally, Section~\ref{sec:dyn} explains how agent interactions can be modeled as dynamic obstacles.

\begin{figure}
\centering
\begin{tikzpicture}[scale=0.8, tips=proper]
    \node[anchor=north] (agent1) at (-3.0, 1.8) {\includegraphics[width=9mm]{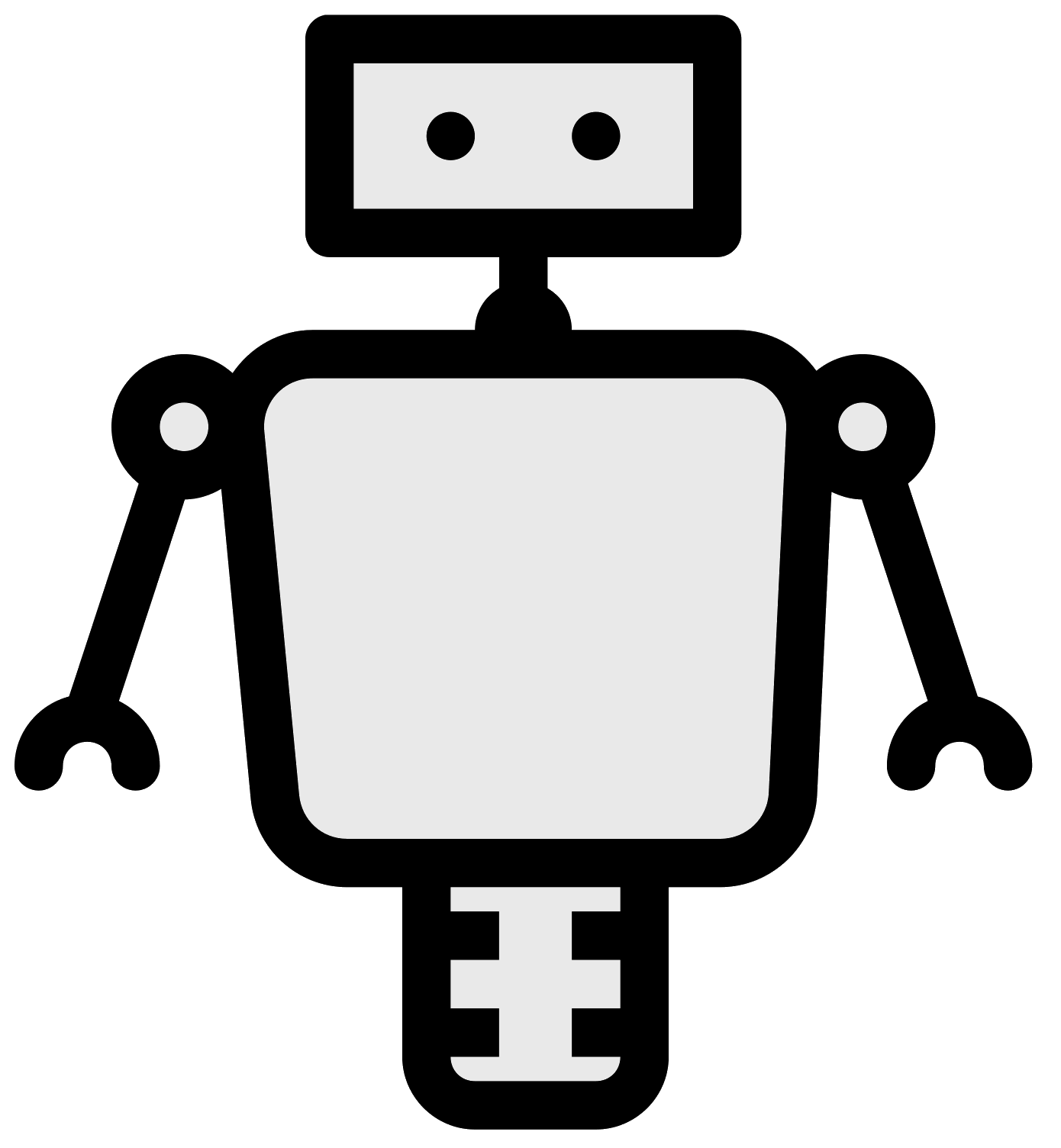}};
    \node[anchor=north] (agent2) at (-3.0, -0.6) {\includegraphics[width=9mm]{image/robot_grey.png}};
    \node[] at (-1.8, 1.1) {Agent$_2$};
    \node[] at (-1.8, -1.4) {Agent$_i$};
    \draw[dashed] (agent1) -- (agent2);

    \node[draw] at (3.0, 0) (nmpc) {NMPC};
    \node[draw] at (1.0, 2) (lio) {Odometry};
    \node[draw] at (-1.0, 2.0) (octo) {Map};

    \node[draw, fill=cyan] at (-1.0, 0.0) (aspt) {A$^*_+$T};
    \node[draw,align=center] at (1, 0) (ptp) {Path\\follower};

    \draw[<->] (-2.9,0.0)  edge node[above]  {$\Psi$} (aspt);
    \draw[->] (aspt) edge node[above]  {$\Psi$} (ptp) ;
    \draw[->] (ptp) edge node[above] {$\psi$} (nmpc) ;

    \draw[->] (octo) edge node[above=3mm, right] {$\mathbf{M}$} (aspt);
    \draw[->] (lio) edge node[above=2mm,right] {$\hat{x}$} (ptp);
    \draw[->] (1.0,1.0) -| (-0.8, 0.3);
    \draw[->] (1.0,1.0) -| (nmpc);

\end{tikzpicture}
\caption{A block diagram of the system, It shows how A$^*_+$T shares its path with the other agents, as well how it interact and depend on other modules. The block map and Odometry can be changed and different options was utilized in this article.}
\label{fig:artchetecture}
\end{figure}

Figure~\ref{fig:artchetecture} illustrates the framework for evaluating A$^*_+$T. While A$^*_+$T is the central focus, ancillary components are also essential. The map block consists of a grid map created offline from an occupancy map via OctoMap~\cite{hornung2013octomap}, converted to a two-dimensional grid map $\mathbf{M}$. In Gazebo simulations, the gmapping technique~\cite{ratul2021design} is used with ground truth odometry, while the odometry source in the experiments is LIO-SAM~\cite{liosam2020shan}. After path planning, A$^*_+$T generates a navigational solution $\Psi$, which is processed by a path follower that integrates a Non-Linear Model Predictive Controller (NMPC)~\cite{karlsson2022ensuring} to produce velocity commands for moving towards waypoints. The main difference between real-world and Gazebo experiments lies in the tuning of components post-execution of A$^*_+$T.

\subsection{A$^*$} \label{sec:A}
The A$^*$ algorithm~\ref{algo:as}~\cite{hart1968formal} is a well-known method for finding the "lowest cost" path $\Psi$ from a robot's current position $\hat{x}$ to a terminal point $\mathbf{B}$ in a grid map $\mathbf{M}$ structured as an undirected graph $G(V,E)$. This "lowest cost" refers to a composite cost function that accounts for both distance and various risk factors, ensuring that the path is safe for traversal, similar to the D$^*_+$ algorithm.

\floatname{algorithm}{Algorithm}
\begin{algorithm}
\caption{A$^*$\\
    \textbf{Input:} A graph G(V,E), starting vertex $\hat{x}$, and end vertex $\mathbf{B}$\\
    \textbf{Output:} $\Psi$
}
\label{algo:as}
    \scalebox{0.6}{
    \begin{minipage}{1.2\linewidth}
\begin{algorithmic}[1]
\STATE \textbf{struct} vertex($v,g,h$):
    \STATE \hspace{0.5cm} $V$ = $v$ \textit{Pointing to a $V$ in G(V,E)} 
    \STATE \hspace{0.5cm} $G = g$ \textit{Cost from vertex to $\hat{x}$}
    \STATE \hspace{0.5cm} $H = h$ \textit{Estimated cost from vertex to $\mathbf{B}$}
    \STATE \hspace{0.5cm} $F = G + H$
    \STATE \hspace{0.5cm} $V_{prevus} =$ NaN \textit{The $V$ arriving from}

\STATE \textbf{procedure} initialize():
    \STATE \hspace{0.5cm} vertex\_sertch = $\{(\hat{x},0,\text{heuristic($\hat{x},\mathbf{B}$)}\}$
    \STATE \hspace{0.5cm} vertex\_added = {$\hat{x}$}
    
\STATE \textbf{procedure} heuristic($a,b$):
    \STATE \hspace{0.5cm} \textbf{return} distance($a,b$)

\STATE \textbf{procedure} cost($a,b$):
    \STATE \hspace{0.5cm} \textbf{return}  distance($a,b$)

\STATE \textbf{procedure} find\_next():
    \STATE \hspace{0.5cm} $i = 0$
    \STATE \hspace{0.5cm} next $=$ vetrex\_sertch$[i]$
    \STATE \hspace{0.5cm} \textbf{while} next.V $\exists$ vertex\_added \textbf{do}
        \STATE \hspace{1.0cm} next $=$ vetrex\_sertch$[i]$
        \STATE \hspace{1.0cm} \textbf{if} i $>$ size(vetrex\_sertch) \textbf{do}
            \STATE \hspace{1.5cm} \textbf{EXIT} \textit{no valid $\Psi$ exists}
        \STATE \hspace{1.0cm} \textbf{end if}
    \STATE \hspace{1.0cm} \textbf{end while}
    \STATE \hspace{0.5cm} \textbf{return} next

\STATE \textbf{procedure} generate\_$\Psi$():
    \STATE \hspace{0.5cm} $\Psi = \emptyset$
    \STATE \hspace{0.5cm} active = vetrex\_sertch$[\mathbf{B}]$
    \STATE \hspace{0.5cm} \textbf{while} active $\neq \hat{x}$ \textbf{do}
        \STATE \hspace{1.0cm} $\Psi$.add(active) \textit{add to front}
        \STATE \hspace{1.0cm} active $=$ active.$V_{prevus}$
    \STATE \hspace{0.5cm} \textbf{end while}
    \STATE \hspace{0.5cm} \textbf{return} $\Psi$

\STATE \textbf{procedure} planner():
    \STATE \hspace{0.5cm} initialize()
    \STATE \hspace{0.5cm} active = vetrex\_sertch$[0]$
    \STATE \hspace{0.5cm} \textbf{while} active.$V \neq \mathbf{B}$ \textbf{do}
        \STATE \hspace{1.0cm} \textbf{for all} $v$ connected $G$ from atrive.$V$ \textbf{do}
            \STATE \hspace{1.5cm} $new = vertex(v, \text{active}.G + \text{cost(v,active.V)}, $\\ \hspace{2.0cm}$ \text{heuristic($v,\mathbf{B}$)})$
            \STATE \hspace{1.5cm} \textbf{if} $v \nexists$ vertex\_added \textbf{DO}
                \STATE \hspace{2.0cm} \textbf{if} $new.G \neq \infty$ \textbf{do}
                    \STATE \hspace{2.5cm} vertex\_sertch.add(new)
                \STATE \hspace{2.0cm} \textbf{end if}
                \STATE \hspace{2.0cm} vertex\_added.add($v$)
             \STATE \hspace{1.5cm} \textbf{else} \textbf{do}
                \STATE \hspace{2.0cm} \textbf{if} $new.F <$ vetrex\_sertch$[new.V].F$ \textbf{do}
                    \STATE \hspace{2.5cm} vetrex\_sertch$[new.V] = new$ 
                \STATE \hspace{2.0cm} \textbf{end if}
            \STATE \hspace{1.5cm} \textbf{end if}     
        \STATE \hspace{1.0cm} \textbf{end for all}
        \STATE \hspace{1.0cm} sort(vetrex\_sertch) \textit{ based on F desending}
        \STATE \hspace{1.0cm} active = find\_next()
    \STATE \hspace{0.5cm} \textbf{end while}
    \STATE \hspace {0.5cm} generate\_$\Psi$()

\end{algorithmic}
\end{minipage}%
    }
\end{algorithm}

The risks considered in this study include occupancy risk ($R_c$), proximity risk ($R_p$), dynamic object risk ($R_o$), time risk ($R_t$), and distance risk ($R_d$). Occupied areas in the grid map are deemed impassable and assigned an infinite risk value.

The total risk is represented as $R = R_c + R_p + R_o + R_t + R_d$. The goal of the A$^*$ algorithm is to minimize the total cost, given by $\min H = \sum{\forall R \in \Psi}$. It uses a heuristic best-first search strategy, focusing on paths with the smallest total cost $F(v) = H(v) + G(v)$, where $H(v)$ estimates the cost to the endpoint, and $G(v)$ reflects the minimum remaining cost. The heuristic $H(v)$ incorporates the risks, defined as follows:

\begin{equation}
G(v) = \begin{cases}
\sqrt{(v_x - \mathbf{B}_x)^2 +(v_y - \mathbf{B}_y)^2} & \text{if } R_c \neq 50 \\
50 \sqrt{(v_x - \mathbf{B}_x)^2 +(v_y - \mathbf{B}_y)^2} & \text{if } R_c = 50
\end{cases}
\label{eq:h}
\end{equation}

Here, $R_c = 50$ indicates uncharted areas. This heuristic enhances efficiency in unknown environments but may overlook optimal paths with smaller unknown segments. This risk for unknown areas also addresses the challenges posed by imperfect gird maps that often have non-existing holes in walls by encourage path in known to be free spaces.
Furthermore, it distinguishes the A$^*_+$T variant from the traditional A$^*$ algorithm, integrating risk factors into the planning process (as shown in algorithm~\ref{algo:cost}).

\floatname{algorithm}{Algorithm}
\begin{algorithm}
\caption{Occupancy cost $R_c$\\
The procedure changed and added procedure for algorithm~\ref{algo:as} to incorporate $R_c$ and heuristics changes connected to it.
}
\label{algo:cost}
    \scalebox{0.6}{
    \begin{minipage}{1.2\linewidth}
\begin{algorithmic}[1]
  
\STATE \textbf{procedure} $R_c(a)$:
    \STATE \hspace{0.5cm} \textbf{if} $G(a) ==$ occupied \textbf{do}
        \STATE \hspace{1.0cm} \textbf{return} $\infty$
    \STATE \hspace{0.5cm} \textbf{else if} $G(a) ==$ free \textbf{do}
        \STATE \hspace{1.0cm} \textbf{return} $0$
    \STATE \hspace{0.5cm} \textbf{else} \textbf{do}
        \STATE \hspace{1.0cm} \textbf{return} $50$
    \STATE \hspace{0.5cm} \textbf{end if}
    
\STATE \textbf{procedure} cost($a,b$):
    \STATE \hspace{0.5cm} \textbf{return} $\cdots + R_c(a)$

\STATE \textbf{procedure} heuristic($a,b$):
    \STATE \hspace{0.5cm} \textbf{return} EQ~\ref{eq:h}

\end{algorithmic}
    \end{minipage}%
    }
\end{algorithm}

\subsection{Proximity risk} \label{sec:prox}
Proximity risk ($R_p$) measures the risk linked to proximity to walls and static objects in map $\mathbf{M}$. To calculate $R_p$ at a vertex, the distance ($d$) to the nearest occupied cell is evaluated within the Region of Interest (ROI). If $d$ exceeds ROI, $R_p$ is zero; if $d$ is below the threshold $ROI_{crit}$, then $R_p$ is $\infty$ (also for occupied vertices). The relationship between distance and $R_p$ is given by:

\begin{equation}
R_p = 99 - (d - 1) \times \left(\frac{98}{ROI}\right)
\label{eq:prox}
\end{equation}

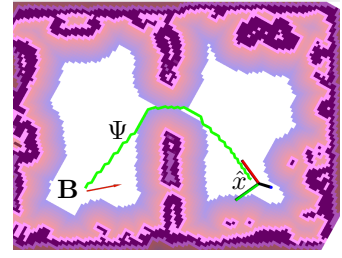
\begin{figure}
    \centering
    \begin{tikzpicture}[scale=0.6]
        \node at (0,0) {\includegraphics[width=0.5\columnwidth]{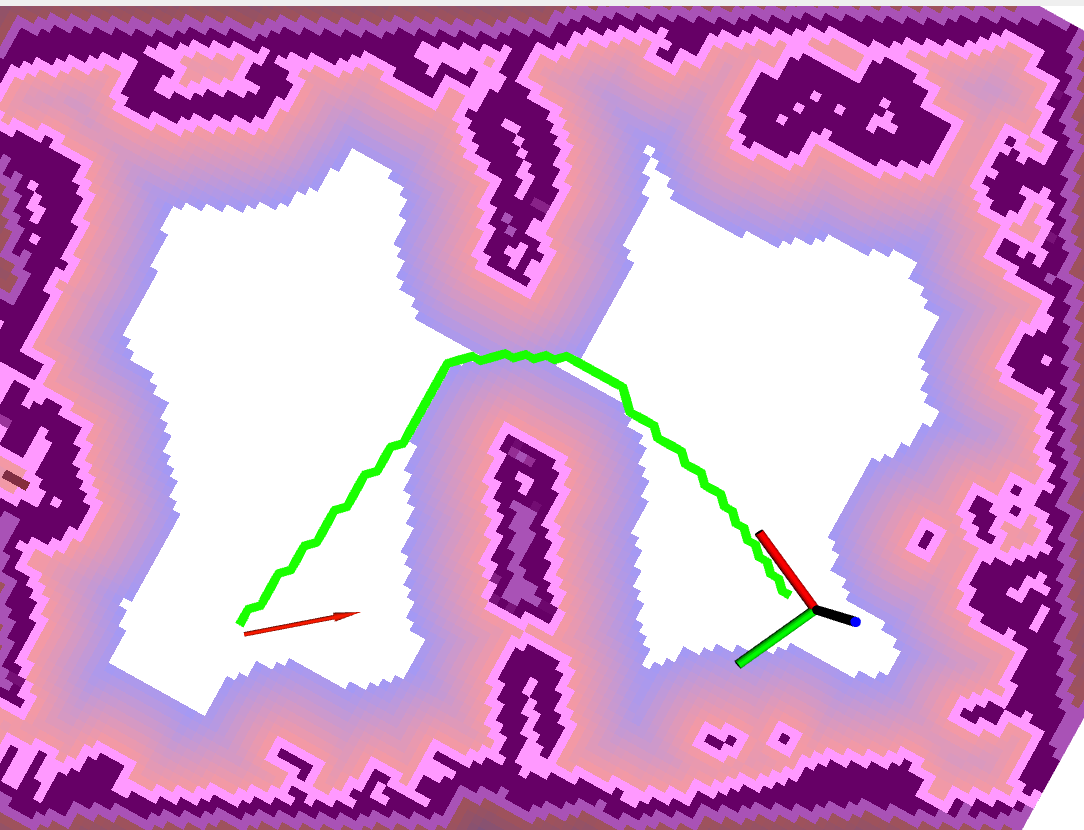}};

        \node[] at (1.4,-1.2) {$\hat{x}$};
        \node[] at (-1.3,-0.1) {$\Psi$};
        \node[] at (-2.4,-1.4) {$\mathbf{B}$};
        
    \end{tikzpicture}
    \caption{An instance where the A$^*_+$T algorithm determines a longer but safer path (green line), opting for a larger opening over a smaller, more hazardous route.}
    \label{fig:risk:layer}
\end{figure}

This ensures $0 < R_p < 100$, creating a risk gradient. Figure~\ref{fig:risk:layer} illustrates this gradient, with purple for occupied areas and light blue for low-risk zones. The impact on trajectory planning shows paths primarily in white regions, indicating free space. The experimental setup and the algorithm adjustments for A$^*_+T$ are detailed in Figures~\ref{fig:lab} and~\ref{algo:prox}, respectively.

\floatname{algorithm}{Algorithm}
\begin{algorithm}
\caption{Proximity risk $R_p$\\
The procedure changed and added form algorithm~\ref{algo:as} to incoperate $R_p$
}
\label{algo:prox}
    \scalebox{0.6}{
    \begin{minipage}{1.2\linewidth}
\begin{algorithmic}[1]
  
\STATE \textbf{procedure} $R_p(a)$:
\STATE \hspace{0.5cm} $p = 0$
    \STATE \hspace{0.5cm} \textbf{for all} ($v \in G$ where distance($v$,$a$) $<$ ROI) \textbf{do}
        \STATE \hspace{1.0cm} \textbf{if} ($G(a) ==$ occupied) \textbf{do}
            \STATE \hspace{1.5cm} \textbf{if} ($p <$ EQ~\ref{eq:prox}(distance($v$,$a$))) \textbf{do}
                \STATE \hspace{2.0cm} $p =$ EQ~\ref{eq:prox}(distance($v$,$a$))
            \STATE \hspace{1.5cm} \textbf{end if}
        \STATE \hspace{1.0cm} \textbf{end if}
    \STATE \hspace{0.5cm} \textbf{end for}
    \STATE \hspace{0.5cm} \textbf{return} $p$

\STATE \textbf{procedure} cost($a,b$):
    \STATE \hspace{0.5cm} \textbf{return} $\cdots + R_p(a)$

\end{algorithmic}
    \end{minipage}%
    }
\end{algorithm}

\subsection{Time} \label{sec:t}

In scenarios where a robot's velocity is known, predicting its position along a path at a future timestamp $k+n\mid k$ is feasible. By aligning $n$ with the grid vertex size and the robot's speed, the robot can effectively traverse one cell per $n$, simplifying future location predictions. This time consideration is integrated into the A$^*$ algorithm with algorithm~\ref{algo:t}.

\floatname{algorithm}{Algorithm}
\begin{algorithm}
\caption{Time\\
The procedure changed and added procedure for algorithm~\ref{algo:as} to incorporate time
}
\label{algo:t}
    \scalebox{0.6}{
    \begin{minipage}{1.6\linewidth}
\begin{algorithmic}[1]

  \STATE \textbf{struct} vertex($v,g,h,n$):
    \STATE \hspace{0.5cm} $\cdots$
    \STATE \hspace{0.5cm} $N = n$ \textit{Time step from planing inissiated}

\STATE \textbf{procedure} cost($a,b$):
    \STATE \hspace{0.5cm} \textbf{return} $\cdots + a.N$

\STATE \textbf{procedure} planner():
    \STATE \hspace{1.5cm} $\cdots$
    \STATE \textit{exhange line 36 in algorithm~\ref{algo:as}}
    \STATE \hspace{1.5cm} $new = vertex(v, \text{active}.H + \text{cost(v,active.V)}, $\\ \hspace{2.0cm}$ \text{heuristic($v,\mathbf{B}$)}, \text{active}.N + 1)$
    \STATE \hspace{1.5cm} $\cdots$

\end{algorithmic}
    \end{minipage}%
    }
\end{algorithm}

If the speed deviates from this framework, future position estimation requires additional calculations, as explained in Section~\ref{sec:pred}. This setup also involves time-varying costs for traversing the graph $G$, where dynamic entities incur temporal costs affecting nearby planning.

By sharing pre-planned trajectories among agents, future locations can be estimated, allowing for collision-free path planning, similar to Conflict-Based Search (CBS) methods. Instead of using collision trees, the ego agent employs A$^*_+$T to create a trajectory that accommodates other agents with established routes.

Traditional A$^*$ algorithms primarily check individual vertex traversability, which can lead to suboptimal paths when dynamic obstacles are present. In many cases, pausing to let a dynamic entity pass may be better than a detour.

Integrating waiting as a maneuver into the A$^*$ algorithm can complicate path planning, necessitating a termination criterion (a watchdog) to avoid infinite waits. The waiting behavior is incorporated into the A$^*$ algorithm~\ref{algo:as} by algorithm~\ref{algo:wait}.

\floatname{algorithm}{Algorithm}
\begin{algorithm}
\caption{Waiting\\
The procedure changes for algorithm~\ref{algo:as} to enable waiting
}
\label{algo:wait}
    \scalebox{0.6}{
    \begin{minipage}{1.2\linewidth}
\begin{algorithmic}[1]

  \STATE \textbf{struct} vertex($v,g,h,n$):
    \STATE \hspace{0.5cm} $\cdots$
    \STATE \hspace{0.5cm} $N = n$ \textit{Time step from planing inissted}
    \STATE \hspace{0.5cm} $w = 0$ \textit{Number of time steps to wait}
    
\STATE \textbf{procedure} cost($a,b$):
    \STATE \hspace{0.5cm} \textbf{if} $R_o \neq 0$ \textbf{do}
        \STATE \hspace{1.0cm} $b.w += 1$
        \STATE \hspace{1.0cm} $b.F += 1$
    \STATE \hspace{0.5cm} \textbf{end if}
    \STATE \hspace{0.5cm} $\cdots$

\STATE \textbf{procedure} generate\_$\Psi$():
    \STATE \hspace{0.5cm} $\Psi = \emptyset$
    \STATE \hspace{0.5cm} active = vetrex\_sertch$[\mathbf{B}]$
    \STATE \hspace{0.5cm} \textbf{while} active $\neq \hat{x}$ \textbf{do}
        \STATE \hspace{1.0cm} \textbf{do}
            \STATE \hspace{1.5cm} $\Psi$.add(active) \textit{add to front}
            \STATE \hspace{1.5cm} active.w -= 1
        \STATE \hspace{1.0cm} \textbf{while} active.w $ > 0$
        \STATE \hspace{1.0cm} active $=$ active.$V_{prevus}$
    \STATE \hspace{0.5cm} \textbf{end while}
    \STATE \hspace{0.5cm} \textbf{return} $\Psi$

\STATE \textbf{procedure} planner():
    \STATE $\cdots$
    \STATE \textit{insert from line 50 in algorithm~\ref{algo:as}}
    \STATE \hspace{1.0cm} \textbf{if} watch\_dog\_timeout() \textbf{do}
        \STATE \hspace{1.5cm} \textbf{EXIT} \textit{solution took to long}
    \STATE \hspace{1.0cm} \textbf{end if}
    \STATE $\cdots$

\end{algorithmic}
    \end{minipage}%
    }
\end{algorithm}

\subsection{dynamic object risk} \label{sec:dyn}

When the speed of a dynamic object doesn’t align with the size of its vertex, calculating its future position over a time step, denoted as $n$, is essential. This allows for determining the object's trajectory.

In dynamic obstacle detection and tracking, as outlined by DTAA~\cite{nordstrom2024dtaa}, obstacle information can be shared among multiple agents as illustrated by figure~\ref{fig:share}. This method, however, can result in a race condition during path planning. The first robot to plan a path, $\Psi$, operates on a clear navigational map, while subsequent robots face a more crowded environment.

\begin{figure}[t]
    \centering
    \input{tikz/share_1}
    \caption{When an agent generates a path $\Psi$, it transmits its $\Psi$ to all other agents for consideration during their path planning. Conversely, when any other agent computes its own $\Psi_i$, it also shares $\Psi_i$ with all other agents, including the initial agent that transmitted a $\Psi$.}
    \label{fig:share}
\end{figure}

When multiple robots plan their $\Psi$ concurrently, they may unintentionally share paths without coordination. To counteract this, robots are programmed to replan their $\Psi$ at regular intervals. This allows them to correct any errors from previous race conditions, addressing inaccuracies in velocity and path tracking. Each new $\Psi$ is evaluated from the robot’s current position, independent of prior discrepancies.

Each dynamic object carries a risk, $R_o$, calculated as follows:

\begin{equation}
 R_o = 99 - \frac{98 d}{ROI + \Upsilon_v \tan(0.44) n} 
 \label{eq:ro}
\end{equation}

Like the risk $R_p$, $R_o$ ranges from $0 < R_o < 100$. Here, $ROI$ has a prediction uncertainty of $25^\circ$ $(\unit[0.44]{rad})$, scaling with the object's velocity $\Upsilon_v$ and the future time interval $n$. The dynamic risk modifies the A$^*$ algorithm as shown in algorithm~\ref{algo:dyn}.

\floatname{algorithm}{Algorithm}
\begin{algorithm}
\caption{Dynamic obstacles $R_o$\\
The procedure changed and added procedure for algorithm~\ref{algo:as} to incorporate dynamic obstacle\\
$\bar{\Upsilon}$ is a list of all dynamic obstacle $\Upsilon_{x,y,\text{major},\text{minor},\theta,\Psi}$
}
\label{algo:dyn}
    \scalebox{0.6}{
    \begin{minipage}{1.2\linewidth}
\begin{algorithmic}[1]

\STATE \textbf{procedure} $R_o$($a$):
    \STATE \hspace{0.5cm} \textbf{for all} ($\Upsilon$ in $\bar{\Upsilon}$ at $k+a.N|k$) \textbf{do}
        \STATE \hspace{1.0cm} \textbf{if} ($\Upsilon_{\text{major}} == \Upsilon_{\text{minor}}$) \textbf{do}
            \STATE \hspace{1.5cm} $d_{local} =$ EQ~\ref{eq:cirkle}($\Upsilon$,x)
        \STATE \hspace{1.0cm} \textbf{else do}
            \STATE \hspace{1.5cm} $\lambda = 0$
            \STATE \hspace{1.5cm} \textbf{while} ($\lambda < ROI$) \textbf{do}
                \STATE \hspace{2.0cm} \textbf{if} (EQ~\ref{eq:colition}($\Upsilon,x$)) \textbf{do}
                    \STATE \hspace{2.5cm} $d_{local} = \lambda$
                    \STATE \hspace{2.5cm} brake
                \STATE \hspace{2.0cm} \textbf{end if}
                \STATE \hspace{2.0cm} $\lambda += 0.1$, $\Upsilon_{\text{major}} += \lambda, \Upsilon_{\text{minor}} += \lambda$
            \STATE \hspace{1.5cm} \textbf{end while}
        \STATE \hspace{1.0cm} \textbf{end if}
        \STATE \hspace{1.0cm} \textbf{if} ($d_{local} < d$) \textbf{do}
            \STATE \hspace{1.5cm} $d = d_{local}$
        \STATE \hspace{1.0cm} \textbf{end if}
    \STATE \hspace{0.5cm} \textbf{end for}
    \STATE \hspace{0.5cm} \textbf{if} $d < ROI_{crit}$ \textbf{do}
         \STATE \hspace{1.0cm} \textbf{return} $\infty$
    \STATE \hspace{0.5cm} \textbf{else} \textbf{do}
        \STATE \hspace{1.0cm} \textbf{return} EQ~\ref{eq:ro}($d, a.N$)
    \STATE \hspace{0.5cm} \textbf{end if}

\STATE \textbf{procedure} cost($a,b$):
    \STATE \hspace{0.5cm} \textbf{return} $\cdots + R_o(a)$

\end{algorithmic}
    \end{minipage}%
    }
\end{algorithm}

\subsubsection{Bounding shape} 

Objects are often represented as circles, but for elongated shapes like the Boston Dynamics Spot robot, an ellipse is a better choice. 

The main difference between a circle and an ellipse is that a circle has one radius, while an ellipse has two axes: the major axis and the minor axis. These axes can be of different lengths and orientations. For circles, determining the distance to the edge is simple:

\begin{equation}
    d = \sqrt{(\Upsilon_x - \hat{x}_x)^2 + (\Upsilon_y - \hat{x}_y)^2} - \Upsilon_{\text{major}}
\label{eq:cirkle}
\end{equation}

In contrast, determining how far a point lies from an ellipse is more complex, typically requiring brute-force methods due to a fourth-degree polynomial. Given parameters $\Upsilon_{x, y, \text{major}, \text{minor}, \theta}$, can it be tested if a point is inside an ellipse with:

\begin{align}
\begin{split}
  & \Upsilon_{\text{major}}^2 \Upsilon_{\text{minor}}^2 - \\
  & \Upsilon_{\text{major}}^2 (\sin \Upsilon_{\theta})^2 - \Upsilon_{\text{minor}}(\sin \Upsilon_{\theta})^2
   (x_{x} - E_{x,k+n\mid k})^2 \\
  & + 2 (\Upsilon_{\text{minor}}^2 - \Upsilon_{\text{major}}^2) \sin (\Upsilon_{\theta}) 
  \cos (\Upsilon_{\theta}) (x_{x} - \Upsilon_{x,k+n\mid k})\\
  &(x_{y} - \Upsilon_{y,k+n\mid k}) + (\Upsilon_{\text{major}}^2 (\cos \Upsilon_{\theta})^2 + \Upsilon_{\text{minor}}^2 (\sin \Upsilon_{\theta})^2) \\
  &(x_{y} - \Upsilon_{y,k+n\mid k})^2 \leq 0
\end{split}
\label{eq:colition}
\end{align}

By slightly increasing the axes $\Upsilon_{\text{major}} $ and $\Upsilon_{\text{minor}}$, we can find the minimum size needed for the ellipse to enclose a point $x$, with this increase representing the distance $d$.

\subsubsection{Prediction}\label{sec:pred}
In path planning, it's crucial to predict the positions of other agents involved in similar processes, especially for dynamic objects that don't reveal their trajectories. Using the Detect Track and Avoid Architecture (DTAA), we estimate the velocities of tracked obstacles, which allows us to predict their future positions based on these velocities. This method facilitates a more adaptive response to the movements of nearby dynamic entities.

\subsection{DTAA}
The Detect Track and Avoid Architecture (DTAA)~\cite{nordstrom2024dtaa} offers an effective solution for detecting and tracking dynamic obstacles in various environments. It identifies obstacles and predicts their future positions under a constant velocity assumption, while clustering nearby objects for efficient management. DTAA has been tested in real-world scenarios with non-communicative dynamic objects, like humans not aligning their movement with the grid.

For a deeper understanding, readers should refer to the original publication, but a brief summary is as follows: DTAA utilizes the YOLOv8 model~\cite{jocher2023yolo} to detect humans in RGB images. These individuals are tracked and spatially mapped in conjunction with depth imagery. A Kalman filter estimates their velocities, aiding in trajectory predictions.

When humans are within a defined safety distance, they are clustered, and this information, along with estimated paths, is shared among agents, allowing them to treat these clusters as dynamic obstacles to avoid.

\section{Presentation of comparison algorithms}
\subsection{CBS}
Conflict-Based Search (CBS) is an optimal and complete algorithm for finding conflict-free paths for multiple agents. It operates on two levels: a high-level that constructs a conflict tree (CT) and performs a best-first search to ensure optimality, and a low-level that determines individual agent paths based on constraints. The CT is a binary tree where each node contains constraints for agents, a solution with their paths, and the total cost.

The root node starts with no constraints. CBS selects the lowest-cost node to identify and resolve conflicts by creating two successor nodes that inherit the parent constraints and add new ones to prevent the conflict. The low level computes optimal paths for each agent based on the constraints. This process continues until a goal node is reached, yielding conflict-free paths for all agents.

\subsection{ECBS}
Focal search~\cite{cohen2018anytime} is a bounded-suboptimal algorithm that uses two node lists: OPEN, the standard open list of $A^*$, and FOCAL, a subset of OPEN where $f$-values are at most $\omega$ times the minimum $f$-value in OPEN. This method allows for alternative heuristics, ensuring that the solution is $\omega$-bounded suboptimal, demonstrated in experiments with $\omega=2$.

Enhanced Conflict-Based Search~\cite{barer2014suboptimal} (ECBS) builds on CBS by integrating focal search at both levels. It maintains a FOCAL list to speed up suboptimal solution identification, guaranteeing solutions no worse than $\omega$ times the optimal. At the low level, ECBS uses the number of conflicts as a heuristic to favor paths with fewer conflicts. The high-level FOCAL list includes nodes with costs below $\omega \cdot \min LB(n)$, where $LB(n)$ is the sum of minimum $f$-values for agents, allowing flexible node expansion while upholding bounded suboptimality.

\subsection{SIPP}

Safe Interval Path Planning~\cite{phillips2011sipp} (SIPP) is a collision-free path planning method based on $A^*$ for a single agent, assuming known trajectories of dynamic obstacles. SIPP identifies safe intervals—contiguous periods where a configuration is free of collisions—and uses "wait and move" actions to generate successors efficiently while avoiding collisions.

SIPP's advantage over time-space $A^*$ lies in its ability to compress wait actions, thereby reducing the number of search nodes. When expanding a node, SIPP generates successors only for actions that lead to neighboring configurations within their safe intervals. This makes SIPP faster in scenarios requiring many wait actions for collision avoidance.

For multiple agents, SIPP uses a priority order where each agent plans its path based on the safe intervals created by the previous agent's path.

\section{Results}

A$^*_+$T was evaluated in simulations within Gazebo across two different environments, compared against CBS, ECBS, and SIPP in five scenarios. To validate its efficacy further, the algorithm was also tested on a Spot robot in lab and corridor settings.
\subsection{Simulation}

To assess the path planning algorithms, a dual-PC setup was used: one PC ran the Gazebo simulation in real-time, while the second executed the path planning algorithm, generating reference paths for the controllers. Both PCs were connected via a wired network. The simulation employed TurtleBot3 Burger robots, controlled using a Nonlinear Model Predictive Control (NMPC) approach, as detailed in~\cite{karlsson2022ensuring}. Prior to planning, the map $\mathbf{M}$ was created using the ROS tool, map-server, and gmapping.

All robot goal poses were provided simultaneously to ensure consistent conditions and minimize variability, preventing A$^*_+$T from gaining an unfair advantage over centralized solutions during planning.

\subsubsection{House}

In a simulated household environment with three robots, the goal is to enable a positional swap through two experimental setups: clockwise and counter-clockwise movements.

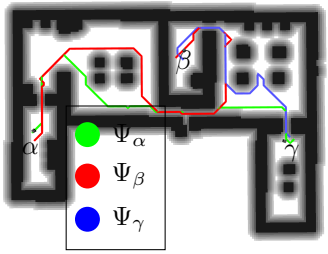
\begin{figure}
    \centering
    \begin{tikzpicture}[scale=0.6,
        greennode/.style={shape=circle, fill=green, line width=2},
        rednode/.style={shape=circle, fill=red, line width=2},
        yellownode/.style={shape=circle, fill=blue, line width=2}
    ]
        \node at (0,0) {\includegraphics[width=0.5\columnwidth]{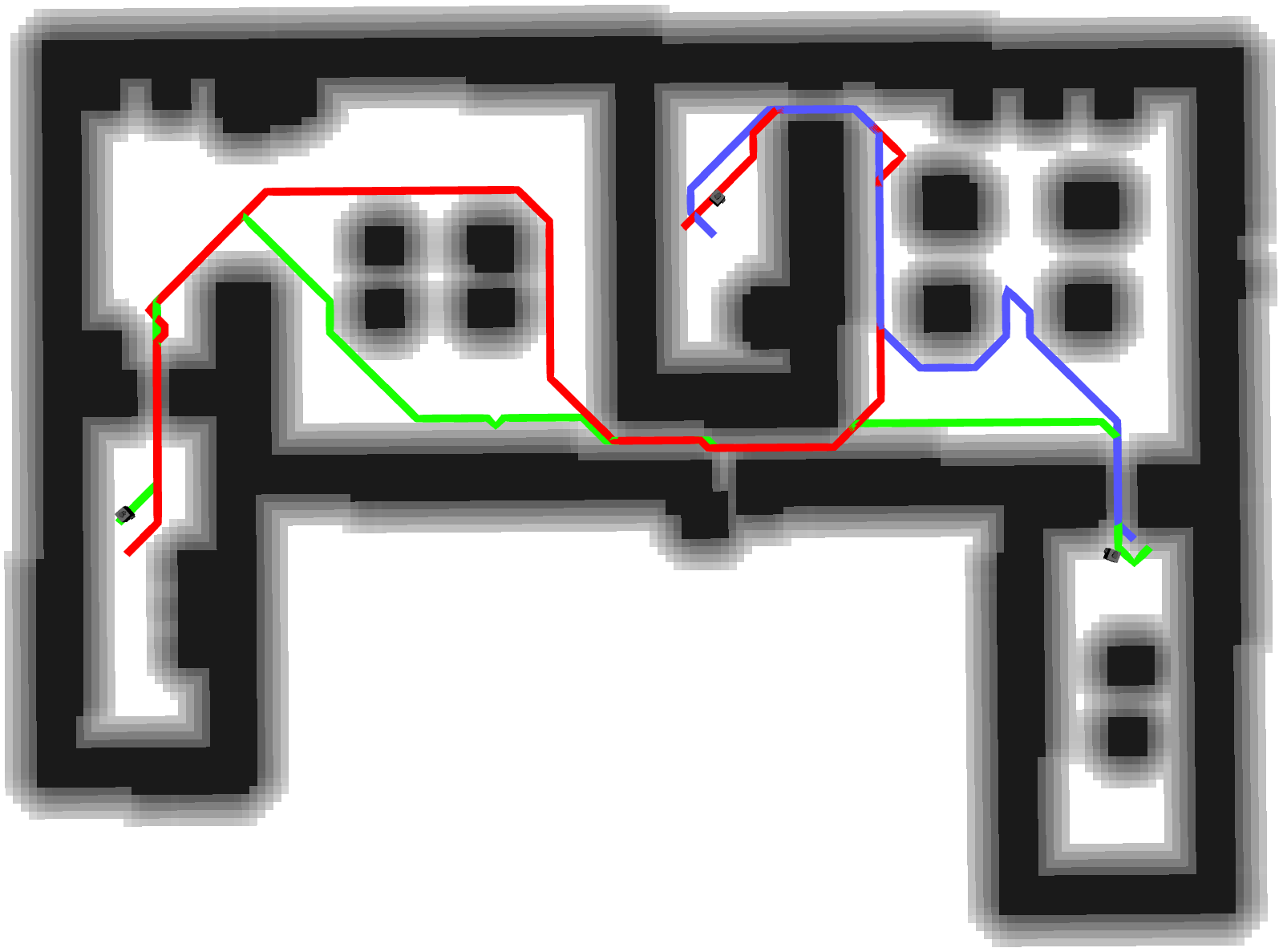}};

        \node[] at (-3.0,-0.6) {$\alpha$};
        \node[] at (0.4,1.3) {$\beta$};
        \node[] at (2.8,-0.7) {$\gamma$};
        
        \matrix [draw, above left] at (current bounding box.south) {
              \node [greennode,label=right:$\Psi_\alpha$] {}; \\
              \node [rednode,label=right:$\Psi_\beta$] {};\\
              \node [yellownode,label=right:$\Psi_\gamma$] {}; \\
            };
    \end{tikzpicture}
    \caption{Three agents plan their trajectory in a simulated house using A$^*_+$T. The map visulized is with the risk layer $R_p$. }
    \label{fig:house:aspt}
\end{figure}

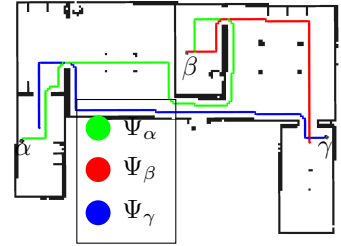
\begin{figure}
    \centering
    \begin{tikzpicture}[scale=0.6,
        greennode/.style={shape=circle, fill=green, line width=2},
        rednode/.style={shape=circle, fill=red, line width=2},
        yellownode/.style={shape=circle, fill=blue, line width=2}
    ]
        \node at (0,0) {\includegraphics[width=0.5\columnwidth]{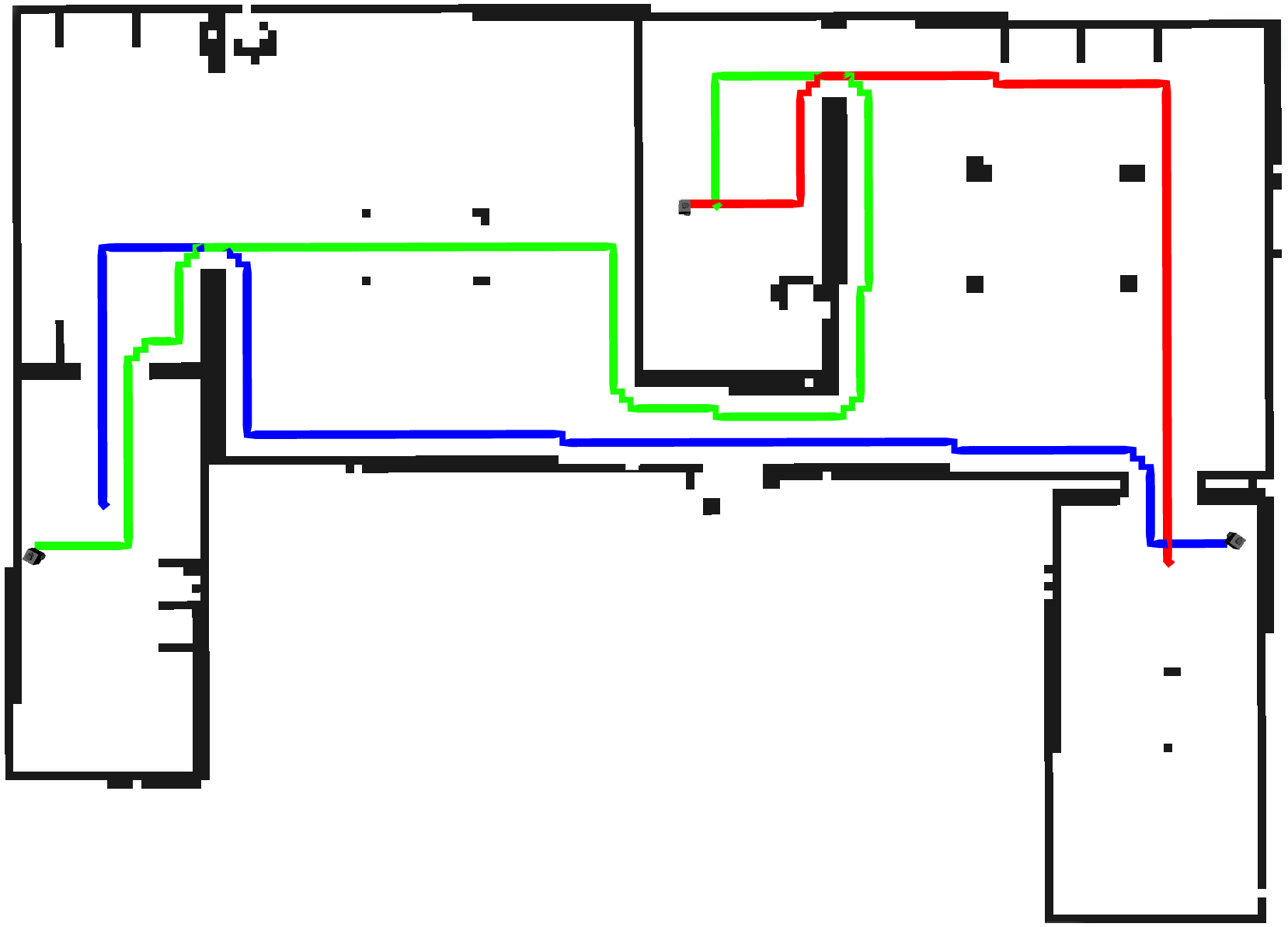}};

        \node[] at (-3.4,-0.7) {$\alpha$};
        \node[] at (0.3,1.1) {$\beta$};
        \node[] at (3.3,-0.7) {$\gamma$};
        
        \matrix [draw, above left] at (current bounding box.south) {
              \node [greennode,label=right:$\Psi_\alpha$] {}; \\
              \node [rednode,label=right:$\Psi_\beta$] {};\\
              \node [yellownode,label=right:$\Psi_\gamma$] {}; \\
            };
    \end{tikzpicture}
    \caption{The SIPP path planed in the house experiment. The path are following the wall with the maximum inflation ($\unit{0.3}{m}$) that where the planning is successful.}
    \label{fig:house:sipp}
\end{figure}

Figure~\ref{fig:house:aspt} illustrate the trajectories generated by the A$^*_+$T algorithm shortly after the robots begin moving with proximity risk ($R_p$) is shown in grayscale. In figure~\ref{fig:house:sipp} is paths from the SIPP algorithm shown, consistent with those from ECBS and CBS, all maintaining proximity to the wall with a safety inflation distance of $\unit{0.3}{m}$. However, with this inflation parameter, the robots struggle to navigate without collisions. Larger safety margins lead to planning failures, highlighting the need for a careful balance between efficiency and safety in robotic navigation in constrained environments.

In Table~\ref{tab:house} the numerical performance measurements from the house simulation are presented.

\begin{table}
\caption{Results for house simulations}
\label{tab:house}
\centering
\resizebox{\columnwidth}{!}{
\begin{tabular}{c | c c c c c }
House clockwise & avg CPU (\%) & max CPU usage (\%)  &memory (Mb) usage & time (s)& total path length (m)\\
\hline
A$^*_+$T  & 47.0 & 55.4 &  3.5 & 2.15 & 57.27\\
CBS& 101.1 & 148.2 &  3.0 & 0.62 & 55.41\\
ECBS  & 85.9 & 105.5 & 3.0 & 0.84 & 55.41\\
SIPP  & 82.5 & 109.1 &  2.0 & 0.44 & 55.50\\
\hline
House  counter-clockwise & & & & &\\
\hline
A$^*_+$T & 35.0 & 45.4 &  3.5 & 2.16 & 56.53\\ 
CBS  & 117.0 & 171.6 &  3.1 & 0.32 & 55.41\\
ECBS  & 90.6 & 147.3 &  3.0 & 0.63 & 55.41\\
SIPP  & 85.9 & 106.9 &  2.0 & 0.39 & 55.50
\end{tabular}
}
\end{table}
\subsubsection{Cave world}
In an expansive cave~\cite{akoval2020}, simulations were conducted to evaluate multi-robot collaboration at a junction with pathways labeled ``a'', ``b'', ``c'', and ``d''. 

\begin{figure}
    \centering
    \begin{tikzpicture}[scale=0.7]
        \node at (0,0) {\includegraphics[width=0.6\columnwidth]{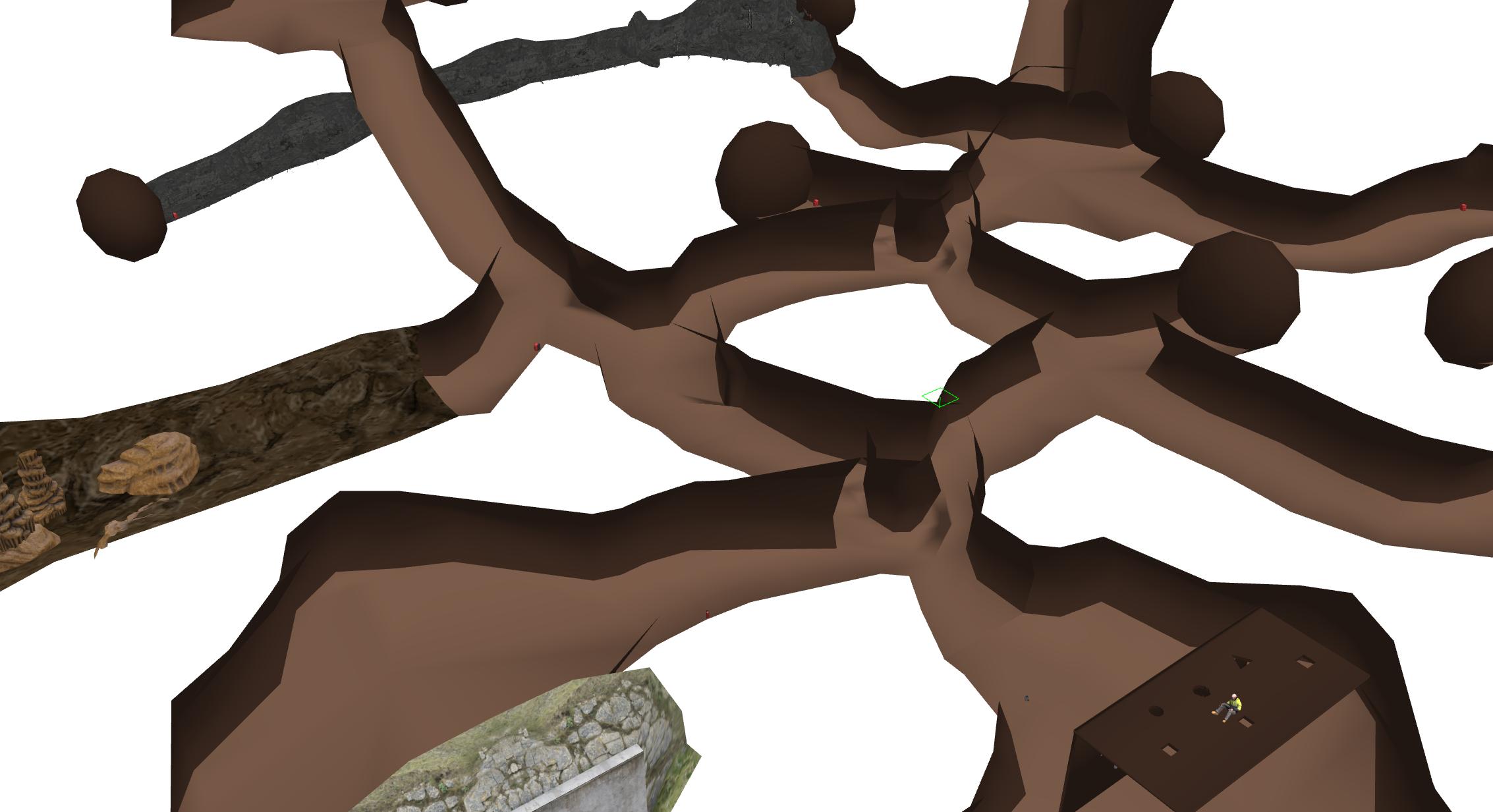}};
        \node[white] at (-0.4 , 0.0) {a};
        \node[white] at (1.4 , -0.0) {b};
        \node[white] at (-0.6 , -1.0) {c};
        \node[white] at (1.4 , -1.3) {d};

    \end{tikzpicture}
    \caption{The gazebo subterranean world used in the experiments.}
    \label{fig:subt}
\end{figure}
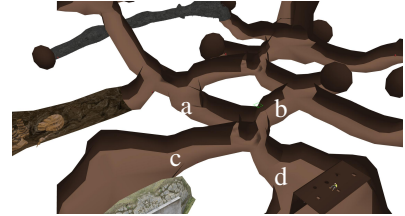

\begin{figure}
    \centering
    \begin{tikzpicture}[scale=0.6,
        greennode/.style={shape=circle, fill=green, line width=2},
        rednode/.style={shape=circle, fill=red, line width=2},
        tealnode/.style={shape=circle, fill=cyan, line width=2},
        yellownode/.style={shape=circle, fill=blue, line width=2}
    ]
        \node at (0,0) {\includegraphics[width=0.6\columnwidth]{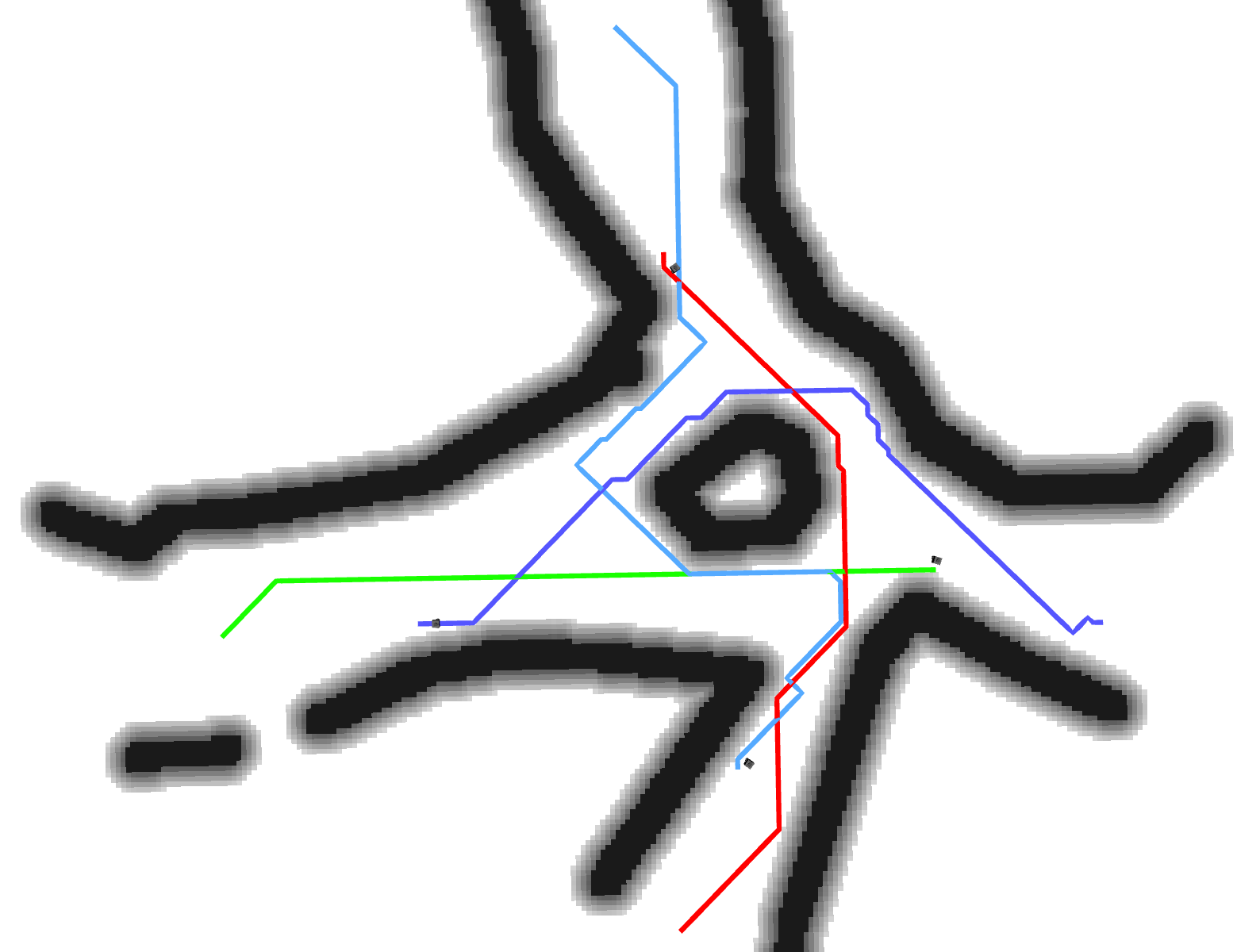}};
        \node[] at (-0.2 , 3.0) {a};
        \node[] at (3.2 , -0.7) {b};
        \node[] at (-2.9 , -0.9) {c};
        \node[] at (0.3 , -2.9) {d};

            \matrix [draw, below right] at (current bounding box.north west) {
              \node [rednode,label=right:$\Psi_a$] {};\\
              \node [greennode,label=right:$\Psi_b$] {}; \\
              \node [yellownode,label=right:$\Psi_c$] {}; \\
              \node [tealnode,label=right:$\Psi_d$] {}; \\
            };
            
    \end{tikzpicture}
    \caption{A snapshot of the paths planned by A$^*_+$T after all agents have moved part of the way. This is from the first scenario when a$\leftrightarrow$d,b$\leftrightarrow$c}
    \label{fig:subt:aspt}
\end{figure}
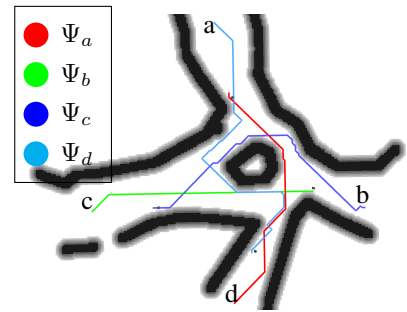

Three sets of simulations were performed, each involving the exchange of robot positions. In the first simulation, robots at ``a'' and ``d'' were swapped, as were those at ``b'' and ``c''. The second involved exchanges between robots at ``a'' and ``b'', and ``c'' and ``d''. In the third, robots at ``a'' and ``c'' were swapped, alongside those at ``b'' and ``d''. These iterations provided insights into the robotic system's dynamics in the cave. Figure~\ref{fig:subt:aspt} displays the first scenario, while Table~\ref{tab:subt} shows the simulation results.

\begin{table}
\caption{Results for subterranean world simulations}
\label{tab:subt}
\centering
\resizebox{\columnwidth}{!}{
\begin{tabular}{c | c c c c c}
Subterranean world first\\ (a$\leftrightarrow$d,b$\leftrightarrow$c) & avg CPU (\%) & max CPU usage (\%)  &memory (Mb) usage & time (s)& total path length (m)\\
\hline
A$^*_+$T & 45.7 & 58.6 &  3.40 & 0.92 & 86.36\\
CBS & 72.7 & 121.5 &  3.2 & 0.93 & 86.17\\
ECBS & 128.2 & 168.4 &  3.0 & 1.15 & 86.17\\
SIPP & 125.1 & 163.9 &  2.1 & 0.37 & 86.17 \\
\hline
Subterranean world second\\ (a$\leftrightarrow$b,c$\leftrightarrow$d)  & & & & & \\
\hline
A$^*_+$T & 44.2 & 49.6 & 3.40 & 6.24 & 77.29\\ 
CBS & 102.1 & 160.4 &  3.2 & 0.76 & 81.37\\
ECBS & 126.8 & 174.3 &  3.0 & 0.48 & 81.37\\
SIPP & 111.7 & 153.6 &  2.1 & 0.44 & 81.37\\
\hline
Subterranean world third\\ (a$\leftrightarrow$c,b$\leftrightarrow$d)  & & & & & \\
\hline
A$^*_+$T & 50.9 & 62.9 &  3.41 & 1.58 & 64.16\\
CBS & 82.2 & 110.6 &  3.1 & 0.45 & 76.57\\
ECBS & 118.4 & 159.6 &  3.1 & 0.55 & 76.57\\
SIPP & 115.7 & 154.6 &  2.1 & 0.40 & 76.57
\end{tabular}
}
\end{table}

\subsection{Experiments}

\begin{figure}
    \centering
    \begin{tikzpicture}[scale=0.6]
        \node at (0,0) {\includegraphics[width=0.5\columnwidth]{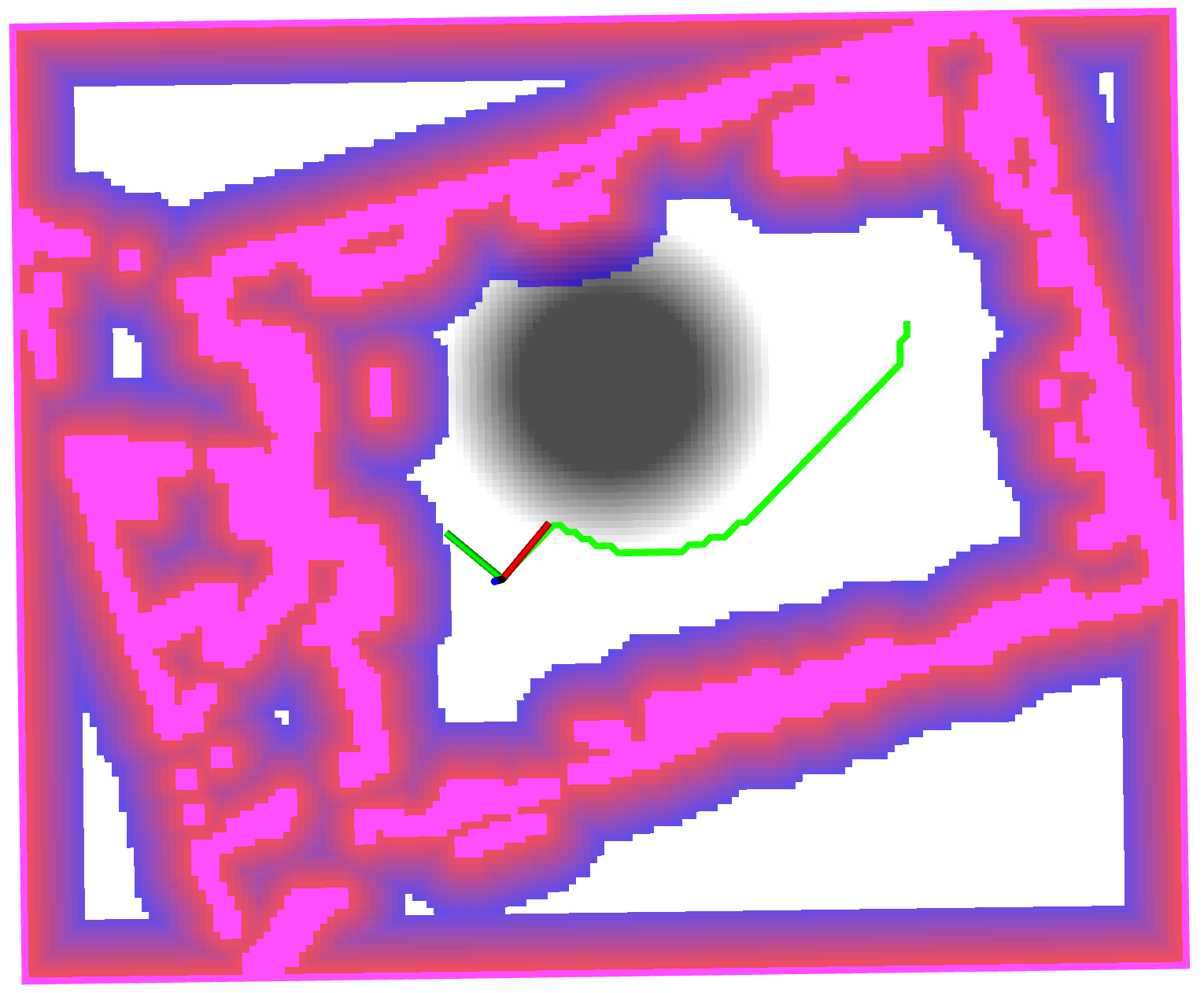}};

        \node[white] at (0.1,0.7) {$R_o$};
        \node[] at (-0.7,-0.8) {$\hat{x}$};
        \node[] at (1.2,0.5) {$\Psi$};
        \node[] at (1.8,1.3) {$\mathbf{B}$};
        
    \end{tikzpicture}
    \caption{A snapshot of a lab experiment where DTAA was used to detect the dynamic obstacle.}
    \label{fig:avoiding}
\end{figure}
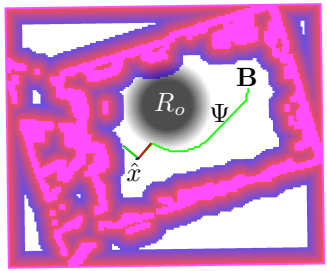

In a laboratory experiment with a Boston Dynamics Spot robot, the A$^*_+$T algorithm was assessed for its ability to navigate dynamic obstacles in real-world environments. As shown in Figure~\ref{fig:avoiding}, a planned path allowed the robot to circumnavigate a dynamic obstacle while ensuring safety margins for human bystanders and walls.

The figure illustrates risk layers, with a static risk gradient in pink and dynamic obstacles in grayscale, highlighting real-time navigation challenges.

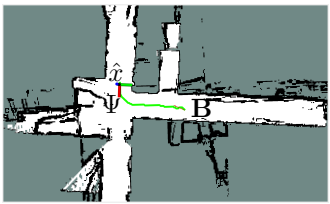
\begin{figure}
    \centering
    \begin{tikzpicture}[scale=0.6]
        \node at (0,0) {\includegraphics[width=0.5\columnwidth]{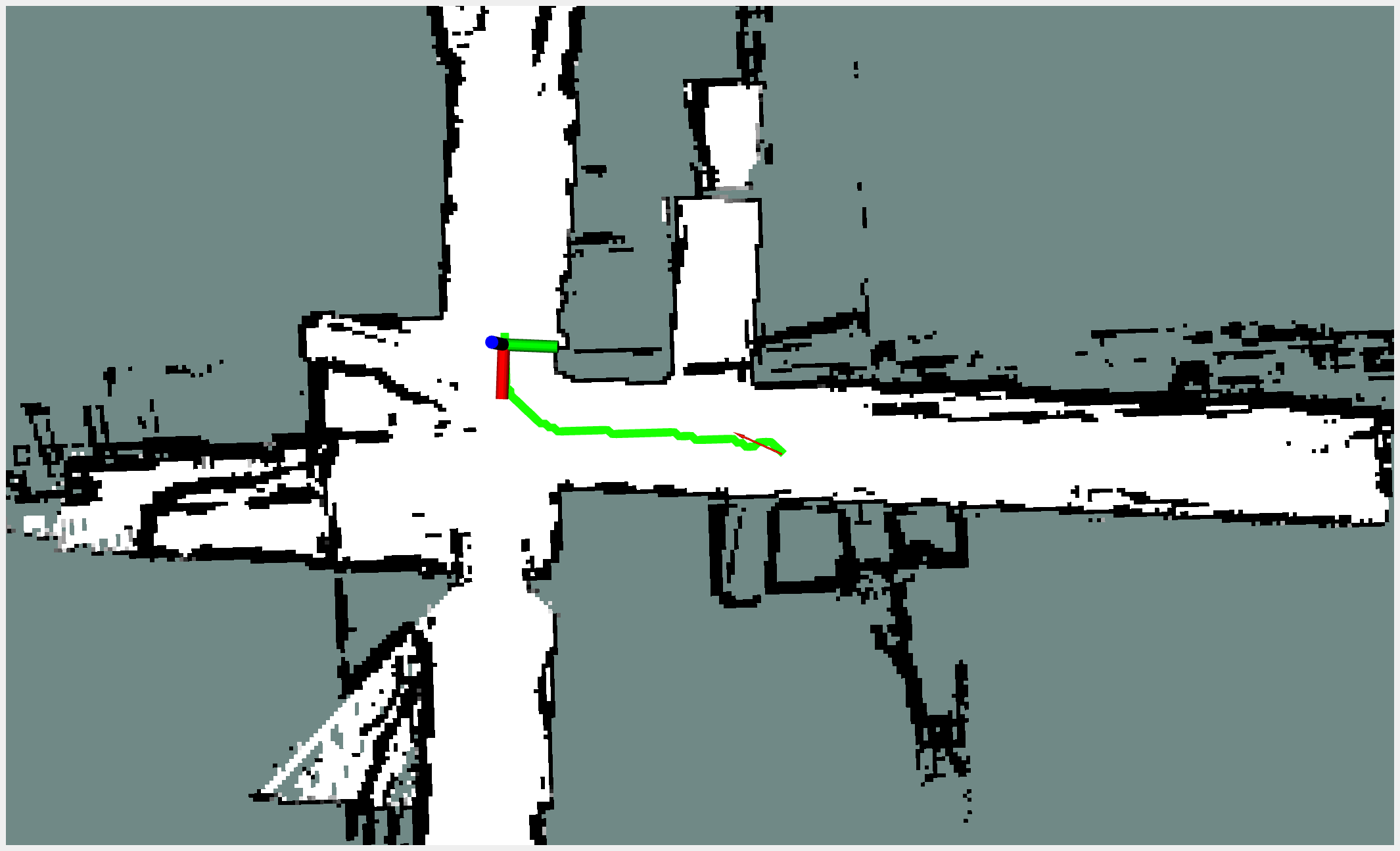}};

        \node[] at (-1.1,0.7) {$\hat{x}$};
        \node[] at (-1.2,0.0) {$\Psi$};
        \node[] at (0.8,-0.1) {$\mathbf{B}$};
        
    \end{tikzpicture}
    \caption{A snapshot of A$^*_+$T planning a path around a corner in a corridor while maintaining an adequate distance to the wall.}
    \label{fig:coridore}
\end{figure}

Further evaluation occurred in a corridor outside the lab, as depicted in Figure~\ref{fig:coridore}, where A$^*_+$T successfully planned a path around a corner while maintaining a safe distance from the walls.

\subsection{Analyses}

When planning robot paths, effective navigation towards a goal is crucial. In the house simulation, the A$^*_+$T planner was the only method tested that successfully generated collision-free paths. Although A$^*_+$T produced slightly longer paths compared to centralized planners due to a larger proximity risk, it was still effective. Centralized planners faced challenges when inflation factors were increased, yielding no viable paths.

Centralized solutions are faster at small scales, but this efficiency may change as the scale increases. A$^*_+$T's distributed operation allowed it to generate valid paths for some agents, as shown in a cave experiment where it offered a timely partial solution, unlike centralized planners which often require all paths to be fully planned before proceeding.

The laboratory experiment also highlighted A$^*_+$T’s ability to consider dynamic objects, setting it apart from other MAPF approaches and demonstrating its adaptability in complex environments.

\section {Conclusions}

In this article, we present the A$^*_+$T algorithm for distributed multi-agent path planning, designed to navigate around dynamic obstacles, including other agents. A$^*_+$T combines a risk strategic approach similar to D$^*_+$ with a time-dependent component to effectively manage these obstacles.

We evaluated A$^*_+$T's performance through multi-agent simulations in Gazebo environments, comparing it to the existing Multi-Agent Pathfinding (MAPF) solutions CBS, ECBS, and SIPP. The results show that A$^*_+$T generates safer and more traversable paths, though it may not find the optimal route.

Additionally, we validated the algorithm in real-world scenarios involving a single robot and a human dynamic obstacle, highlighting A$^*_+$T's practical potential. Future work will focus on testing the algorithm with multiple agents and exploring improvements by adding motion constraints for non-omnidirectional robots.

\bibliographystyle{plain}
\bibliography{IEEEabrv, references.bib}

\end{document}